\documentclass[conference]{IEEEtran}
\IEEEoverridecommandlockouts
\usepackage{cite}
\usepackage{amsmath,amssymb,amsfonts}
\usepackage[ruled,linesnumbered,boxed,noend]{algorithm2e}
\usepackage{graphicx}
\usepackage{textcomp}
\usepackage{xcolor}
\usepackage{booktabs}
\usepackage{tabularx} 
\usepackage{multirow} 
\usepackage[colorlinks, linkcolor=black, anchorcolor=red, citecolor=black]{hyperref}

\usepackage{threeparttable}
\def\BibTeX{{\rm B\kern-.05em{\sc i\kern-.025em b}\kern-.08em
    T\kern-.1667em\lower.7ex\hbox{E}\kern-.125emX}}
\begin{document}

\title{DyPho-SLAM : Real-time Photorealistic SLAM in Dynamic Environments}
\author{
    \IEEEauthorblockN{
     Yi Liu\textsuperscript{1},
     Keyu Fan\textsuperscript{1},
     Bin Lan\textsuperscript{2},
     Houde Liu\textsuperscript{1,*},
    }
    \IEEEauthorblockA{
      \textsuperscript{1}Tsinghua Shenzhen International Graduate School, Tsinghua University, Shenzhen, China \\
      \textsuperscript{2}Jianghuai Advance Technology Center, Hefei, China
    }
    \vspace{-0.3cm}
    \thanks{* Corresponding authors: Houde Liu (Emails: liu.hd@sz.tsinghua.edu.cn) }
    \thanks{This paper was supported by Shenzhen Science Fund for Distinguished Young Scholars under Grant RCJC20210706091946001.}
    
}

\maketitle

\begin{abstract}
Visual SLAM algorithms have been enhanced through the exploration of Gaussian Splatting representations, particularly in generating high-fidelity dense maps. While existing methods perform reliably in static environments, they often encounter camera tracking drift and fuzzy mapping when dealing with the disturbances caused by moving objects. This paper presents DyPho-SLAM, a real-time, resource-efficient visual SLAM system designed to address the challenges of localization and photorealistic mapping in environments with dynamic objects. Specifically, the proposed system integrates prior image information to generate refined masks, effectively minimizing noise from mask misjudgment. Additionally, to enhance constraints for optimization after removing dynamic obstacles, we devise adaptive feature extraction strategies significantly improving the system's resilience. Experiments conducted on publicly dynamic RGB-D datasets demonstrate that the proposed system achieves state-of-the-art performance in camera pose estimation and dense map reconstruction, while operating in real-time in dynamic scenes.
\end{abstract}


\section{Introduction}
\label{sec:Introduction}
Visual simultaneous localization and mapping (vSLAM), which involves determining the pose of a vision sensor and reconstructing the environment, is widely applied in the fields of robotics,  augmented reality (AR), and virtual reality (VR)\cite{1}. Moreover, compared to constructing sparse point clouds, dense reconstruction offers substantial advantages in terms of both accuracy and fidelity\cite{3}. However, in real-world scenes that contain dynamic objects, the traditional dense SLAM algorithms could experience significant performance degradation in terms of incorrect data associations\cite{4}. 

Accomplishing dense SLAM tasks in dynamic environments presents two key challenges: 1) The dynamic objects exclusion strategies are imperfect \cite{7}, leading to local errors in localization. 2) Remaining features are few in dynamic environments, resulting in the reduced constraints for pose optimization \cite{feature} and lack of density in reconstruction.
To address these challenges, this paper proposes a real-time precise localization and high-fidelity mapping system in dynamic scenes. First, we propose a refined mask generation strategy based on prior image information, which comprehensively utilizes the accumulation of past image frames to avoid missing dynamic object masks in a single frame. Next, we develop a feature point selection method with compensation for sparse remaining features, which enhances constraints for optimization. Finally, we integrate mask and feature points into the explicit Gaussian Splatting mapping process, achieving high-fidelity mapping in a dynamic environment.

\begin{figure}[t]
    \centering
    \includegraphics[width=8.5cm]{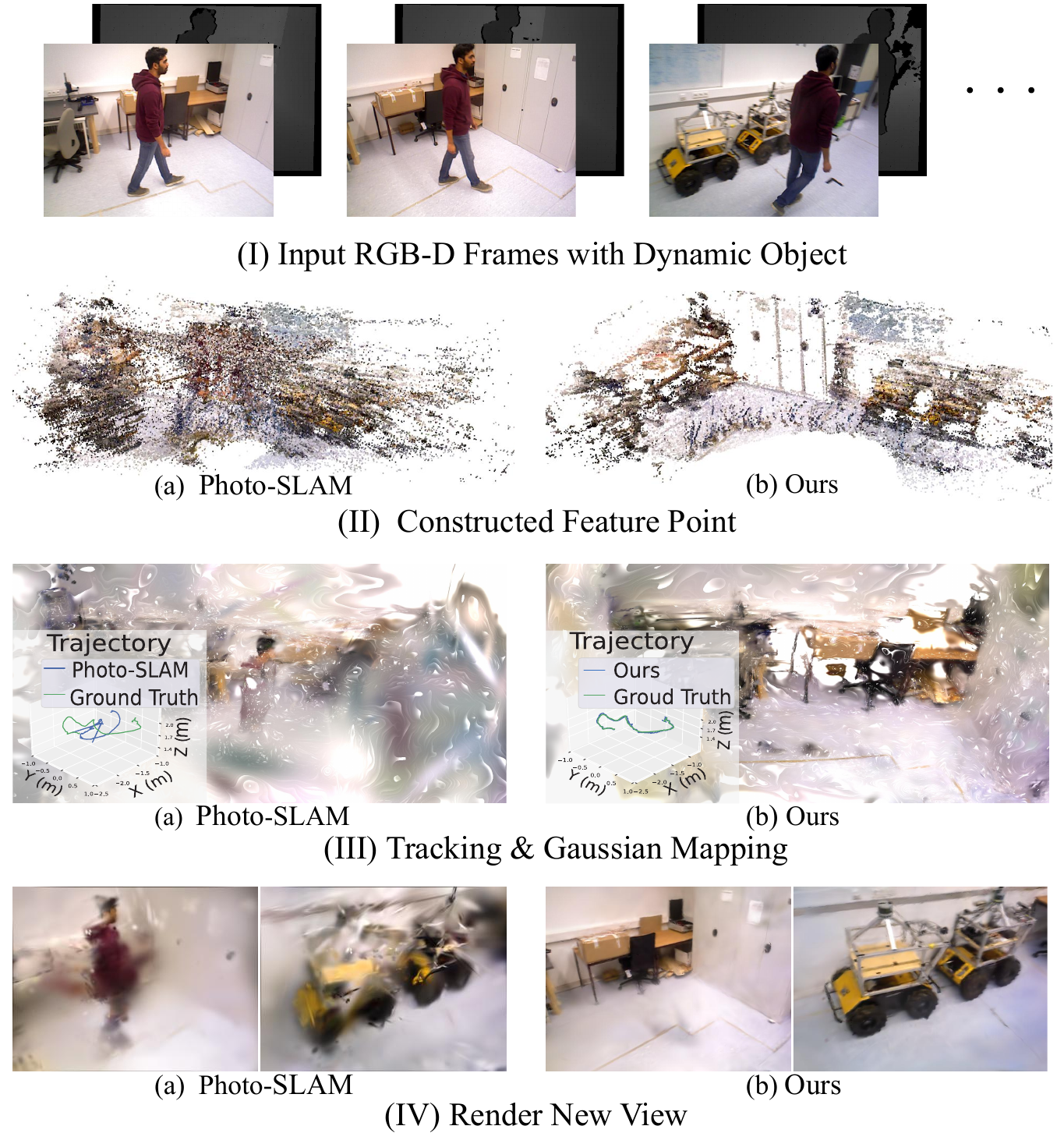}
    \caption{An example compares DyPho-SLAM with Photo-SLAM (Feature-GS) on the bonn/ps\_track dataset. (I) The input RGBD images contain the moving person. (II) presents a comparison of the effectiveness of the constructed feature points, while (III) demonstrates the tracking results and the constructed Gaussian. (IV) illustrates the effect of rendering a new view through Splatting.}
    \label{first}
    \vspace{-0.2cm}
\end{figure}

\begin{figure*}[t]
    \center
    \includegraphics[width=17.5cm]{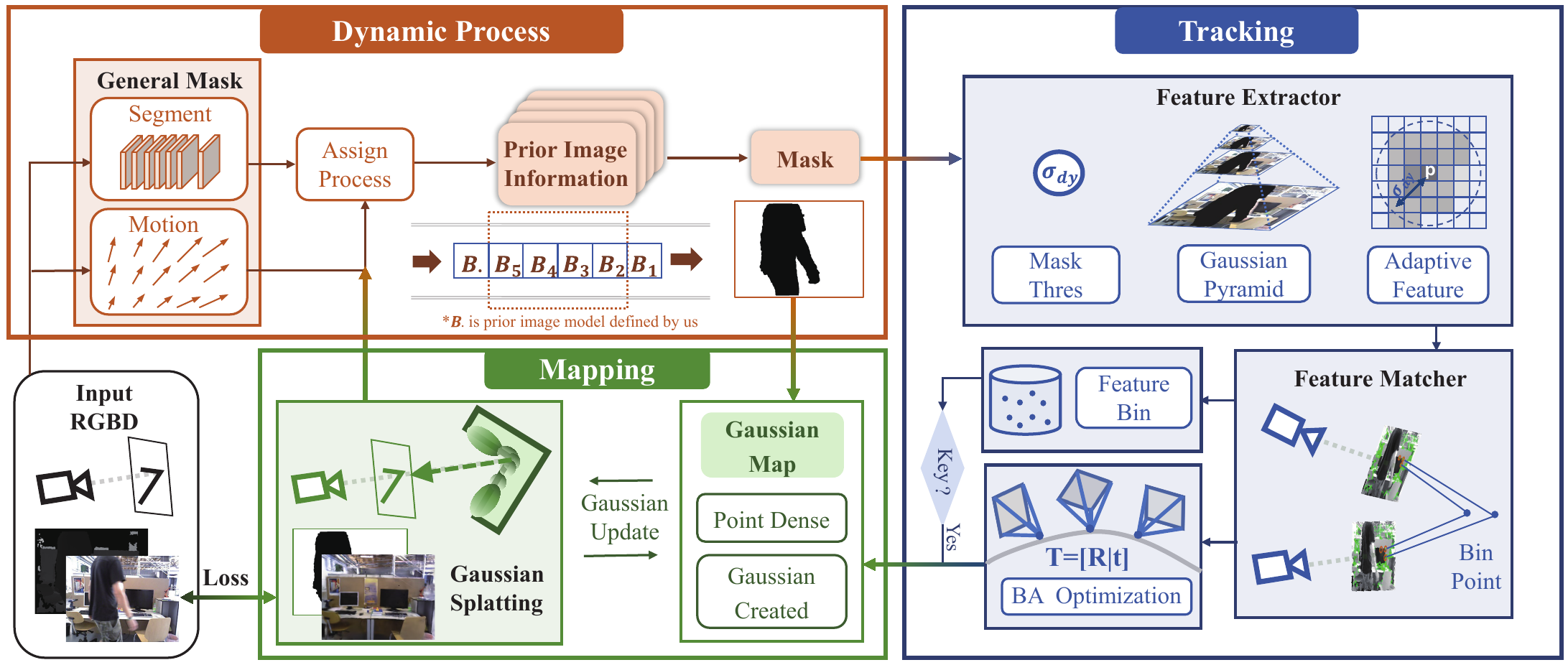}
    \caption{
    Pineline of our Dypho-SLAM: the ``Dynamic Process" module \ref{subsec:NLOS Identification} performs RGB-D image to extract mask, followed by general mask generation and prior image refinement steps. Subsequently, the ``Tracking" module \ref{subsec:Mapping Algorithm} passes through the feature extractor, feature matcher, and pose optimization to obtain the camera pose and feature bin. Finally, the ``Mapping" module \ref{subsec:Location Algorithm}  continuously updates the Gaussian map by optimizing the loss function.}
    \label{Fig:framework}
    \vspace{-0.2cm}
\end{figure*}

\subsection{Related Work}
Current research on dynamic-SLAM primarily focuses on single-frame image detection, neglecting the correlation and continuity between frames, and thus fails to achieve perfect extraction across all frames. For example, ORB-SLAM3 \cite{orbslam} uses re-sampling and residual optimization strategies to remove dynamic objects. Dyna-SLAM \cite{dynaslam} employs semantic extraction to extract dynamic object masks, while Rodyn-SLAM \cite{rodyn-slam} further integrates optical flow motion on this basis. The missed masks in specific frames are then applied to dense mapping methods, which can result in trajectory inaccuracies and some noise artifacts during the rendering process. Our proposed DyPho-SLAM establishes a background model based on previous images, achieving highly accurate mask generation and reducing the noise present in dynamic mapping.\par
Considering the resource consumption issue of implicit representation, 3D Gaussian Splatting (3DGS) framework achieves rapid rendering requirements through explicit scene representations\cite{5}, enabling its rapid development in the field of dense SLAM\cite{10}. MonoGS\cite{gsslam} and SplatAM\cite{splatam} simultaneously optimize tracking and Gaussian parameters to accomplish SLAM tasks, while Photo-SLAM\cite{photoslam} initializes with feature points as hyper-primitives and addresses the speed issues in both tracking and mapping. Currently, dynamic 3DGS SLAM is still in a rapid exploratory stage. The latest dynamic 3DGS SLAM algorithms, DGS-SLAM\cite{dgsslam} and DG-SLAM\cite{dgslam} fail to meet the requirements for real-time and precise localization, resulting in the phenomenon of ghosting artifacts in rendering. 
\subsection{Contributions}
This work offers the following contributions: 
\begin{enumerate}
\item We present an explicit representation-based framework capable of achieving real-time camera tracking and high-quality mapping in real-world dynamic environments.
\item We propose a high-quality mask generation strategy based on prior image information and an adaptive point selection method, resulting in accurate localization and high-fidelity reconstruction.

\item Experiments on two challenging dynamic datasets demonstrate Dypho-SLAM achieves state-of-the-art performance in camera pose estimation and dense map reconstruction, while operating in real-time.
\end{enumerate}

\section{Overview of the Framework}
\label{sec:framework}

\subsection{Problem Statement}
\label{subsec:problem}
\
Given an image sequence $\mathcal{I}=\{\boldsymbol{I}_i|\boldsymbol{I}_i=[\boldsymbol{C}_i|\boldsymbol{D}_i]\}_{i=1}^N (\boldsymbol{I}_i\in \mathbb{R}^{H\times W\times4}, \boldsymbol{C}_i\in\mathbb{R}^{H\times W\times3}, \boldsymbol{D}_i\in\mathbb{R}^{H\times W\times1})$ containing dynamic objects captured at a certain frame rate by an RGB-D camera with known intrinsics $K\in \mathbb{R}^{3\times3} $, our objective is to obtain their corresponding camera projection matrices $\mathcal{T}=\{\boldsymbol{T}_i|\boldsymbol{T}_i=[\boldsymbol{R}_i|\boldsymbol{t}_i]\}_{i=1}^N$, while updating Gaussian parameters $\mathcal{G}=\{\boldsymbol{\mu}_j,\alpha_j,\boldsymbol{\Sigma_j},\boldsymbol{c}_j\}_{j=1}^{n(\mathcal{G})}$in every frame as follows:
\begin{equation}
    \label{equ:exception}
	\begin{aligned}
\mathcal{I}\to\mathcal{T},\mathcal{G}
\end{aligned}
\end{equation}
where the camera orientation $\mathbf{R}\in SO(3)$
and position $\boldsymbol{t}\in \mathbb{R}^3$ together form the camera's projection matrix $\boldsymbol{T}\in SE(3)$. The Gaussian parameter is characterized by its spatial position $\boldsymbol{\mu}_j$, opacity value $\alpha_j$, covariance matrix $\boldsymbol{\Sigma}_j$, and spherical harmonics-based color coefficients $\boldsymbol{c}_j$.
The process of solving $\mathcal{T}$ and $\mathcal{G}$ should satisfy:  
1) minimizing absolute trajectory error root mean square error.
2) optimal Gaussian rendering performance for representing the static environment.
3) minimizing the time and resources needed to complete the task.\par
\subsection{System Framework}
The structure of the proposed Dypho-SLAM algorithm completes real-time localization and photorealistic mapping tasks in dynamic environments through a series of interconnected processes, as shown in Fig. \ref{first} and Fig. \ref{Fig:framework}. 
It primarily comprises three modules: the ``Dynamic Process'' module \ref{subsec:NLOS Identification}, the ``Tracking'' module \ref{subsec:Mapping Algorithm}, and the ``Mapping'' module \ref{subsec:Location Algorithm}.


\section{Implementation}
\label{sec:implementation}

\subsection{Dynamic Process}
\label{subsec:NLOS Identification}

\subsubsection{Raw Mask Generation}
\ 
\newline
\indent 
To distinguish between dynamic obstacles and the static environment, we need to obtain a mask of moving obstacles, uniformly represented as $M_{[*]} (M_{[*]}\in \mathbb{R}^{H\times W\times 1})$, where 0 and 1 denote the presence and absence of a moving obstacle at a pixel, respectively. We obtained the raw mask by overlaying two mask generation methods as follows:\par

\begin{itemize} 
\item[$\bullet$] \textbf{Segment Mask}: 
The method of obtaining a mask through semantic segmentation provides precise segmentation for image semantic understanding as follows:
\begin{equation}
    \label{equ:distribution}
	\begin{aligned}
f_{\boldsymbol{\theta}}:\{\boldsymbol{I}_i\}_{i=1}^N\to\{M_{seg}\}_{i=1}^{N}
	\end{aligned}
\end{equation}
where $f_{\boldsymbol{\theta}}$ is implemented as a YOLO network \cite{yolo} with trainable parameters $\theta$ trained on large-scale datasets.
\item[$\bullet$]  \textbf{Motion Mask}: To mitigate the impact of out-of-distribution moving objects on segment extraction, we introduce the optical flow method. Specifically, we solve for the optical flow vectors using the optical flow constraint equation as follows:
\begin{equation}
    \label{equ:exception}
	\begin{aligned}
\frac{\partial L}{\partial x}u+\frac{\partial L}{\partial y}v+\frac{\partial L}{\partial t}=0
\end{aligned}
\end{equation}
where $L(x,y,i)$ is the i-th image's brightness function, $u$ and $v$ are the components of optical flow vector. After determining optical flow, we apply thresholds  $T$ to identify motion masks on larger magnitudes, as follows:
\begin{equation}
    \label{equ:exception}
	\begin{aligned}
M_{Flow}(x,y)=\left\{\begin{matrix}0&||{u(x,y)+v(x,y)}||^2>T\\1&\text{otherwise}\end{matrix}
\right.
\end{aligned}
\end{equation}
\end{itemize}

\subsubsection{Optimized Mask via Prior Image Information}
\ 
\newline
\indent 
 The semantic and motion methods only focus on a single image frame, ignoring the continuity between frames, which can lead to mask misjudgments and result in significant noise and ghosting in map construction. Here, we propose a method for removing dynamic obstacles based on a prior static background image model, which is specifically represented as follows:
 \begin{equation}
    \label{equ:exception}
	\begin{aligned}
B_i = (1-\tau-\rho)B_{i-1}+\tau R_d(\mathcal{G}, \boldsymbol{T}_{i-1}\delta\boldsymbol{ T})\\+ \rho\cdot \boldsymbol{D}_i(\boldsymbol{1}-M_{Flow}\bigoplus M_{Segment})
\end{aligned}
\end{equation}
where $B_i(B_i\in \mathbb{R}^{H\times W\times 1})$ represents the depth static image model of the i-th image and $B_1$ is initialized by removing the dynamic objects from the first depth image frame. $\tau$ and $\rho$ are hyperparameters, representing the degree of belief in past renderings and the degree of trust in the segment mask and flow motion mask, respectively. $\bigoplus$ is an assign process, which can be simply processed through OR operation. $R_d(\cdot)$ is a Gaussian rendering process to return depth image through a Gaussian map and camera pose detailed as \ref{render}, and $\delta\boldsymbol{T}$ is determined by a constant velocity camera model to estimate a relative pose value as follows:
\begin{equation}
    \label{equ:exception}
	\begin{aligned}
\delta\boldsymbol{ T} = \boldsymbol{T}_{i-1}\boldsymbol{T}_{i-2}^{-1}
\end{aligned}
\end{equation}
Here, we obtain the final mask result by considering a point as static if its neighboring depth value is similar to the value in the static map model. Specifically:
\begin{equation}
    \label{equ:exception}
	\begin{aligned}
M(x,y)=\left\{\begin{matrix}1&n(|B_{i,(x,y)}-\boldsymbol{D}_{i,(N(x,y))}|<\sigma_m)>n_{m}\\0&\text{otherwise}\end{matrix}\right.
\end{aligned}
\end{equation}
where $N(\cdot)$ returns the neighbors of $(x, y)$. $\sigma_m$ measures the depth difference threshold. The function $n(\cdot)$ returns the count of neighbors that satisfy the condition within $(\cdot)$, and $n_{m}$ represents the threshold for this count.

\subsection{Localization Algorithm}
\label{subsec:Mapping Algorithm}
\subsubsection{Feature Extract}
\label{subsec:Grid}
\
\newline
\indent
\begin{figure*}[t]
    \centering
    \includegraphics[width=18cm]{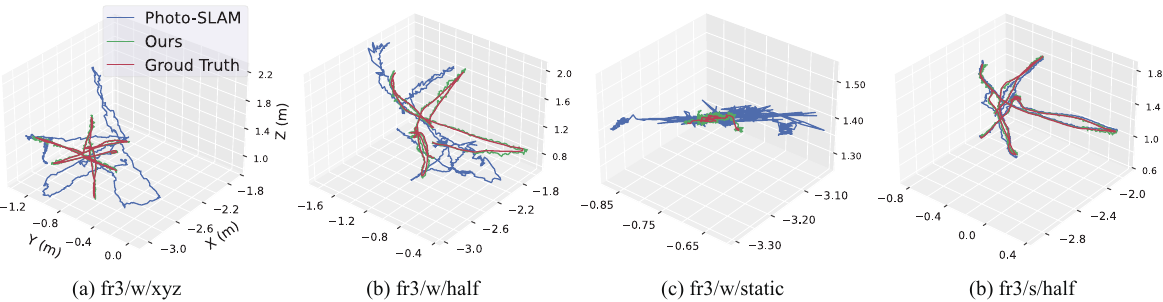}
    \caption{Camera trajectories estimated by Photo-SLAM and ours in the TUM dataset sequences, and the differences from ground truth}
    \label{ate}
    \vspace{-0.3cm}
\end{figure*}
\begin{table*}[t]
\caption{\textbf{Camera tracking results on dynamic scenes from the TUM RGB-D dataset. }The units for ATE$(\downarrow)$ and Std.$(\downarrow)$ are in CM. The best results among all domains are highlighted in \textbf{bold}. $\times$ represents that the corresponding data is not in the original report.}
\centering
\begin{tabular}{cc|cc|cc| cc| cc| cc}
\toprule
\textbf{Method}&\textbf{Dense}&\multicolumn{2}{c|}{\textbf{fr3/w/xyz}}&\multicolumn{2}{c|}{\textbf{fr3/w/half}}&\multicolumn{2}{c|}{\textbf{fr3/w/static}}&\multicolumn{2}{c|}{\textbf{fr3/s/half}}&\multicolumn{2}{c}{\textbf{Avg.}}\\
\midrule
\textit{Feature-based SLAM methods} & & ATE & Std. & ATE & Std. & ATE & Std. & ATE & Std. & ATE & Std. \\
ORB-SLAM3 & & 28.1 & 12.2 & 0.31 & 9.0 & 2.0 & 1.1 & 2.6 & 1.6 & 15.8&6 \\
Dyna-SLAM & & 1.7 & $\times$ & \textbf{2.6} & $\times$& 0.7 & $\times$ & 2.8 & $\times$ & 2&$\times$\\
\midrule
\textit{NeRF-based SLAM methods} & &ATE&Std.&ATE&Std.&ATE&Std.&ATE&Std.&ATE&Std. \\
NICE-SLAM&$\checkmark$ & 113.8 & 48.9 & $\times$ & $\times$ & 137.3 & 21.7 & 93.3 & 35.3 & 114.8 &33.3 \\

ESLAM &$\checkmark$& 45.7& 28.5 & 60.8 & 27.9 & 93.6 & 20.7 & 3.6 & 1.6 & 34.5 &13.5 \\
RoDyn-SLAM &$\checkmark$& 8.3 & 5.5 & 5.6 & 2.8 & 1.7 & 0.9 & 4.4 & 2.2 & 4.1&2.3 \\
\midrule
\textit{3DGS-based SLAM methods} & &ATE&Std.&ATE&Std.&ATE&Std.&ATE&Std.&ATE&Std.\\
Splatam &$\checkmark$& 136.6& 33.7 & 185.4 & 67.6 & 78.3 & 23.5 & 14.1 & 7.0 & 135.5 &109.6 \\
GS-SLAM &$\checkmark$& 37.2& 9.9 & 60.0 & 20.7 & 8.4 & 4.1 & 7.4 & 5.4 & 28.5 &10.0 \\
GassiDy &$\checkmark$& 3.5& 1.6 & 3.7 & 1.9 & \textbf{0.6} & 0.3 & 2.4 & 1.4 & 2.6 &1.3 \\
DGS-SLAM &$\checkmark$& 4.1& 2.2 & 5.5 & 2.8 & \textbf{0.6} & \textbf{0.2} & 4.1 & 1.6 & 3.0 &1.5 \\

\midrule
\textit{Feature-3DGS SLAM methods} & &ATE&Std.&ATE&Std.&ATE&Std.&ATE&Std.&ATE&Std.\\
Photo-SLAM &$\checkmark$& 88.3& 48.9 & 33.5 & 9.3 & 9.4 & 6.3 & 3.2 & 1.4 & 33.6 &16.3 \\
Ours &$\checkmark$& \textbf{1.6}& \textbf{0.8} & \textbf{2.6} & \textbf{1.3} & \textbf{0.6} & 0.3 &\textbf{1.6} & \textbf{0.7} & \textbf{1.6} &\textbf{0.7} \\
\bottomrule
\end{tabular}

\label{tum}
\vspace{-0.3cm}
\end{table*}
After obtaining the mask, we excluded the areas containing dynamic obstacles. However, removing these dynamic obstacles resulted in a decrease in feature points, which reduced the constraints within the SLAM optimization problem, making it underdetermined and degenerate. Consequently, localization performance decreased compared to static environments. To address the issue of fewer feature points in dynamic environments, we propose a dynamic feature sampling method based on a generated mask on the Gaussian pyramid. A Gaussian pyramid represents an image at multiple scales, capturing varying levels of detail through a series of Gaussian smoothing and downsampling operations applied iteratively to the original image. When a pixel satisfies the following criteria, we define it as a relaxed feature points specific to dynamic masking:
\begin{equation}
    \label{equ:motion model}
	\begin{aligned}
n(|\boldsymbol{G}_{i,(x,y)}-\boldsymbol{G}_{i,(N(x,y))}|<\sigma_{dy})>n_{f}
        \end{aligned}
\end{equation}
where $G$ represents the grayscale image of the i-th frame image at pixel $(x, y)$, $n_f$ is a number value above which a point can be considered as a feature point, and $\sigma_{dy}$ is a dynamic threshold used to adjust the number of feature points in dynamic environments, which is expressed as:
\begin{equation}
    \label{equ:motion model}
	\begin{aligned}
\sigma_{dy}=\sigma_{0}(1-k\frac{\sum{M}}{H\times W})
        \end{aligned}
\end{equation}
where $\sigma_0$ represents the initial gray difference threshold, and $k$ is a parameter used to adjust the proportion of additional feature points. This method ensures a stable number of feature points as a bin after applying the mask, thereby achieving a balance between processing speed and localization accuracy.

\subsubsection{Pose Optimization}
\label{Free-Occupied Mapping}
\
\newline
\indent 
In the localization process, we use a motion-only bundle adjustment to minimize the reprojection error between matched 2D geometric keypoint of the frame and 3D point, which we are trying to optimize with LM algorithm\cite{lm} is
\begin{subequations}
  \begin{align}
    \{\boldsymbol{R}_i,\boldsymbol{t}_i\}&=\operatorname{arg\,min}_{\mathbf{R},\mathbf{t}}\sum_{j\in \mathcal{M}}\rho_{h}({\bf e}_{i,j}^{T}{\bf\Omega}_{i,j}^{-1}{\bf e}_{i,j})\\
    wh&ere \quad{\bf e}_{i,j}={\bf P}_{I}-\pi_{i}({\bf T}_{i},{\bf P}_{w,j})
  \end{align}
\end{subequations}
The error term ${\bf e}_{i,j}$ is for the observation of a map point j in i-th image which is optimized by minimizing robust Huber cost function $\rho_{h}(\cdot)$.  $\Omega_{i,j}=\omega_{i,j}^2\bf{I}_{2X2}$ is the scale-associated covariance matrix of the keypoint. $P_w$ represents the coordinates in the world coordinate system, and $P_I$ represents the matched pixel position in the camera coordinate system. $\pi(\cdot)$ is the 3D-to-2D projection function as follows
\begin{equation}
    \label{equ:motion model}
	\begin{aligned}
\pi_{i}(\mathbf{T}_{i},\mathbf{P}_{w,j})=K\boldsymbol{D}_{i,\bf P_w}^{-1} (\boldsymbol{R} {\bf P}_{w,j}+\boldsymbol{t})
        \end{aligned}
\end{equation}
\par
Afterwards, loop closure detection is performed at the keyframe to further optimize the pose.

\subsection{Mapping Algorithm}

We employ the Gaussian map representation, which enables the projection of 3D Gaussian volumes onto a 2D-pixel plane through rasterization rendering, thereby generating images from various viewpoints. The core task in this section is to maintain the Gaussian parameters
$\mathcal{G}=\{\boldsymbol{\mu}_j,\alpha_j,\boldsymbol{\Sigma_j},\boldsymbol{c}_j\}_{j=1}^{n(\mathcal{G})}$.
\label{subsec:Location Algorithm}
\subsubsection{Incrementally Constructed Gaussian Map}
\ 
\newline
\indent 
After obtaining a new keyframe, we incrementally create Gaussians to optimize the map. For newly observed feature point in the image, we add a new Gaussian with the color of that pixel, centered at the location obtained by unprojection function $\pi^{-1}(\cdot)$ to the pixel, an opacity of 0.5, and a unit radius. Furthermore, the 2D geometric feature points essentially represent regions of complex texture within the spatial distribution framework. For complex textures, more Gaussians are required to better fit the semantics. Therefore, we actively create 
additional Gaussians based on the 2D feature points by spatially sampling within the surrounding unit space to 
form new Gaussians and densify the Gaussian map.
\subsubsection{Rendering via Splatting}
\label{render}
\
\newline
\indent 
Gaussian Splatting generates RGB images by first sorting a given set of 3D Gaussians based on their depth from the camera. Subsequently, the image is rendered by alpha-compositing the 2D projections of these Gaussians with $N$ in pixel space, in that order. The final color $\boldsymbol{C}$ and depth $\boldsymbol{D}$ of each pixel is determined by this process.
\begin{equation}
    \label{equ:motion model}
	\begin{aligned}
    \boldsymbol{C}_{r}&=\sum_{i\in N}\mathbf{c}_i\alpha_i\prod_{j=1}^{i-1}(1-\alpha_i)
        \end{aligned}
\end{equation}
Similarly, we can render the depth values of the image,
\begin{subequations}
  \begin{align}
    \boldsymbol{D}_{r}&=\sum_{i\in N}\mathbf{d}_i\alpha_i\prod_{j=1}^{i-1}(1-\alpha_i)\\
    wh&ere \quad\mathbf{d}_i=(T_i\boldsymbol{\mu})_z
  \end{align}
\end{subequations}
\subsubsection{Gaussian Update}
\ 
\newline
\indent 
The Gaussian parameters are then updated iteratively by gradient-based optimization through differentiably rendering RGB, depth, and dynamic mask to minimize the following loss with the Stochastic Gradient Descent algorithm\cite{descent}:
\begin{equation}
    \label{equ:motion model}
	\begin{aligned}
\mathcal{L}=(1-\lambda)\left|M_c\cdot(\boldsymbol{C}_r-\boldsymbol{C}_{gt})\right|_1+\lambda|M\cdot(\boldsymbol{D}_r-\boldsymbol{D}_{gt})|_1
        \end{aligned}
\end{equation}
$\lambda$ is a weight factor for balance color and depth of pixel and $M_c$ represents the repetition of the single-channel $M$ to form a three-channel mask suitable for $\boldsymbol{C}$.

\section{EXPERIMENTS}
\subsection{Experiment Setup }

\begin{figure*}[t]
    \centering
    \includegraphics[width=18cm]{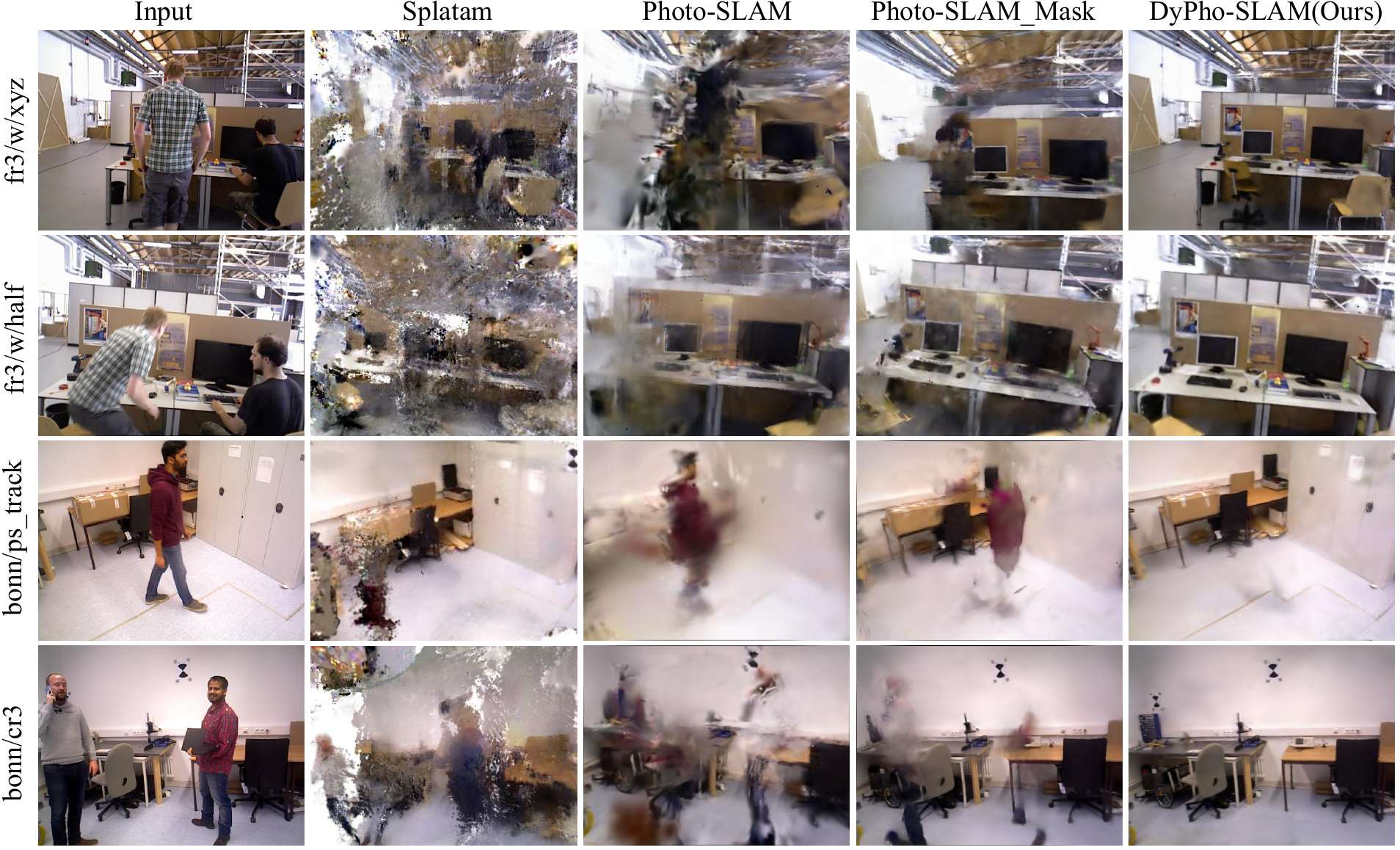}
    \caption{ Comparison of rendered results from state-of-the-art Gaussian Splatting SLAM methods. }
    \label{mapping}
    \vspace{-0.2cm}
\end{figure*}

\textbf{Implementation details}: We adopt Photo-SLAM \cite{photoslam} as the baseline in our experiments and run DyPho-SLAM on a desktop
PC with a 5.20GHz Intel Core i7-13700F CPU and NVIDIA GeForce RTX 3060. DyPho-SLAM was fully implemented in C++ and CUDA, making use of YOLO \cite{yolo}, ORB-SLAM3 \cite{orbslam}, 3D Gaussian splatting \cite{5}, and the LibTorch framework. We set the parameter as follows: $n_f=12,n_m=9,k=0.9,\sigma_m=0.2,\sigma_0=0.3,\lambda=0.7$.\par
\textbf{Datasets and Metric}: We evaluated our approach using two prominent dynamic datasets: the TUM RGB-D dataset \cite{tum} and the Bonn RGB-D dataset \cite{bonn}, which are captured in indoor
environments with a handheld device, providing RGB images, depth maps, and ground truth trajectories. We evaluate camera tracking performance using the Root Mean Square Error (RMSE) and Standard Deviation (Std.) of Absolute Trajectory Error (ATE) \cite{tum}. For the evaluation of reconstruction quality, we qualitatively demonstrated the experimental results through novel view rendering. Additionally, we measure real-time performance by examining the tracking time, mapping time, and processing time per frame.
\begin{table}[t]  
  \centering  
    \caption{\textbf{Ablation study on prior-image-based mask refined strategy and adaptive feature selection method on TUM f3/w/xyz.}
  }  
  \begin{tabular}{c|c|cc}  
    \toprule
    Prior Image Mask& Adaptive Feature & fr3/w/xyz & fr3/w/half  \\  
    \midrule
    \checkmark & $\times$ & 3.12 & 3.83  \\  
    $\times$ & \checkmark & 2.76 & 3.54  \\  
    \checkmark & \checkmark & \textbf{1.62} & \textbf{2.55} \\  
    \bottomrule
   
  \end{tabular}

  \label{ablition}  
   \vspace{-0.2cm}
\end{table}
\subsection{Evaluation of Tracking and Mapping}
\label{subsec: Real-world experiments}
 For a visual demonstration of localization, we compare the trajectories obtained using our method with those from Photo-SLAM on the TUM dataset, as shown in Fig. \ref{ate}. To evaluate the accuracy of camera tracking in dynamic scenes, we compare our methods with traditional SOTA classical SLAM methods like ORB-SLAM3 \cite{orbslam}, DynaSLAM \cite{dynaslam},  dense reconstruction systems like NICE-SLAM \cite{niceslam}, ESLAM \cite{eslam}, Splatam \cite{splatam}, Photo-SLAM \cite{photoslam}, GS-SLAM \cite{gsslam} and recently dynamic SLAM like RoDyn-SLAM \cite{rodyn-slam}, DGS-SLAM \cite{dgsslam}, GassiDy \cite{Gassidy} . As illustrated in Table \ref{tum}, our proposed method attains superior ATE results compared to other dense SLAM approaches and even outperforms SOTA Feature-SLAM in multiple scenarios. In addition, we conducted ablation experiments on the TUM dataset for our proposed method. The specific ATE results, as shown in Table \ref{ablition}, demonstrate the effectiveness of our presented module.\par
To assess the mapping capabilities, we benchmarked our method against  open-source Gaussian Splatting-based SLAM algorithms, notably SplaTAM \cite{splatam} and Photo-SLAM \cite{photoslam}. To ensure fairness, we enhanced the baseline Photo-SLAM with a raw mask specifically for this evaluation. Fig. \ref{mapping} presents a qualitative comparison of the rendered images obtained from the reconstructed Gaussian maps. Traditional GS-based SLAM systems often produce Gaussians for moving objects in such settings, leading to distorted scene reconstructions. In contrast, our proposed methods demonstrate superior performance in generating a precise and photorealistic static map.
\begin{table}[t]  
  \centering  
   \caption{\textbf{Run-time comparison on TUM fr3/w/xyz. }  The best results and second best are in \textbf{bold} and \underline{underlined}, respectively. $\times$ represents the corresponding data isn't in the original report.
  }  
  \begin{tabular}{ccccc}  
    \toprule
     \multirow{2}{*}{Method}& Tracking  & Mapping  & FPS & \multirow{2}{*}{Operate Time $\downarrow$}  \\  
    & [ms] $\downarrow$ & [ms] $\downarrow$ & [hz] $\uparrow$ & \\  
    \midrule
    ORB-SLAM3 & \textbf{52.5} & $\times$ & \textbf{19.2}  & \textbf{$<$1 min} \\  
    Dyna-SLAM3 & 141 & $\times$ & 11.7  & $<$2 mins \\  
    NICE-SLAM & 3535 & 3055 & 0.15  & $>$30 mins \\  
    ESLAM & 1002 & 703 & 0.58 & $>$10 mins\\  
    RoDyn-SLAM & 159 & 678 & 1.19 & $>$10 mins\\  
    Splatam & 3417 & 759  & 0.23 & $>$30 mins\\  
    GS-SLAM & 775 & 164  & 1.07 & $>$10 mins\\  
    Photo-SLAM & \underline{51.5} & \textbf{2.5}  & \underline{17.53} & \underline{$<$1 min 30s}\\  
    DGS-SLAM & $\times$ & $\times$ & 1.60  & $<$5 mins \\  
    DG-SLAM & 89.2 & 549.3 & 1.55 & $<$5 mins\\  
    Ours & 61.7 & \underline{3.2}  & 16.08 & \underline{$<$1 min 30s}\\  
    
    \bottomrule
    
  \end{tabular}
  \label{time}
  \vspace{-0.2cm}
\end{table}

\subsection{Time Analysis}
We evaluated the runtime performance of various SLAM algorithms on the TUM dataset, and the results are presented in Table \ref{time}. For dense SLAM methods, only Photo-SLAM and our proposed DyPho-SLAM achieve frame processing rates that satisfy the requirements for real-time reconstruction. This demonstrates the effectiveness of our framework in addressing dynamic SLAM challenges while utilizing computational resources efficiently.
\section{Conclusion and Future Work}
\label{sec:conclusion}
This study proposes DyPho-SLAM, a real-time and robust method to achieve localization and photorealistic
mapping in dynamic environments.  A novel mask generation method, leveraging prior image information for adaptive feature point selection, has been proposed to address inaccurate dynamic object removal and limited point constraints in high-fidelity mapping. Extensive experiments have demonstrated that DyPho-SLAM significantly outperforms existing SOTA SLAMs for online photorealistic mapping. However, our mapping methods focus on reconstructing a static background, and prior-map-based mask is still constrained by the underlying raw object detection. In the future, we aim to develop a method of reconstructing the motion of dynamic objects in real-time while ensuring the static scene remains stationary.

\bibliographystyle{IEEEbib}
\bibliography{icme2025_template_anonymized}

\vspace{12pt}

\end{document}